\documentclass[10pt,twocolumn,letterpaper]{article}

\usepackage{iccv}
\usepackage{times}
\usepackage{epsfig}
\usepackage{graphicx}
\usepackage{amsmath}
\usepackage{amssymb}
\usepackage{multirow}
\usepackage{graphicx}
\usepackage{subfigure}
\usepackage{color}
\usepackage{booktabs}
\usepackage{tabulary}
\usepackage{rotating}
\usepackage{bbding}

\usepackage[pagebackref=true,breaklinks=true,letterpaper=true,colorlinks,bookmarks=false]{hyperref}

\iccvfinalcopy 

\def\iccvPaperID{666} 

\ificcvfinal\fi
\begin{document}

\title{Face Detection through Scale-Friendly Deep Convolutional Networks}

\author{Shuo Yang\\\and Yuanjun Xiong\\\and Chen Change Loy\\\and Xiaoou Tang\\\and
\small{Department of Information Engineering, The Chinese University of Hong Kong}\\
{\tt\small \{ys014, yjxiong, ccloy, xtang\}@ie.cuhk,edu.hk}
}

\maketitle
\thispagestyle{empty}

\begin{abstract}
In this paper, we share our experience in designing a convolutional network-based face detector that could handle faces of an extremely wide range of scales. 
We show that faces with different scales can be modeled through a specialized set of deep convolutional networks with different structures. These detectors can be seamlessly integrated into a single unified network that can be trained end-to-end.
In contrast to existing deep models that are designed for wide scale range, our network does not require an image pyramid input and the model is of modest complexity.
Our network, dubbed ScaleFace, achieves promising performance on WIDER FACE and FDDB datasets with practical runtime speed. Specifically, our method achieves \textbf{76.4} average precision on the challenging WIDER FACE dataset and $96\%$ recall rate on the FDDB dataset with \textbf{7} frames per second (fps) for $900\times1300$ input image.
\end{abstract}

\section{Introduction}
\label{sec:intro}

Face detection is one of the most studied problems in computer vision~\cite{zhang2010survey}. It serves as a fundamental step to many facial analysis applications, including face alignment, face recognition, and face parsing.
%
%
%
While existing deep learning-based detectors have achieved supreme recognition accuracy, the difficulty of finding faces of a wide range of scales remains.
In many real-world applications such as public space visual surveillance, faces typically appear in different sizes, and they are all required to be detected. 

\begin{figure}[t]
\begin{center}
\includegraphics[width=0.9\linewidth]{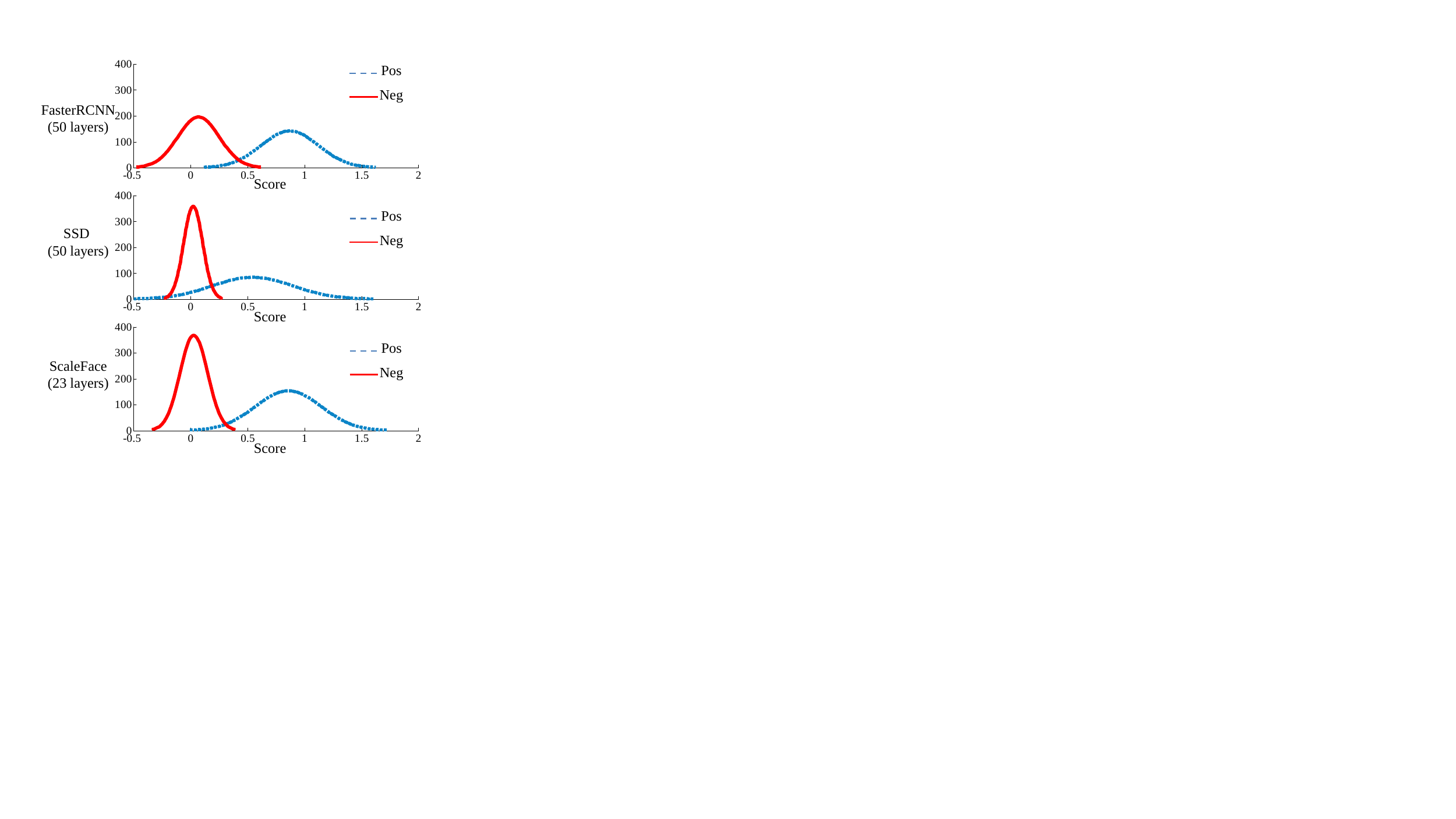}
\caption{\textbf{A small network may be more well-suited for small faces}. The distribution of foreground and background score generated from different methods. Foreground samples come from faces in the range of $10px-40px$. Similarly background are the negative patch in the same range of scale with the foreground.}
\vspace{-0.7cm}
\label{fig:intro}
\end{center}
\end{figure}

Previous methods~\cite{jiang16fasterrcnn,cascadecnn,yang2015faceness} learn highly discriminative scale-invariant representation to solve this problem, which is difficult since the clues to be gleaned for recognizing a $300$-pixels tall face are qualitatively different than those for recognizing a $10$-pixels tall face~\cite{hu2016finding}. 
%
%
Here is the \textit{dilemma} --
%
%
On one hand, more convolution layers are required to learn highly representative features that can distinguish faces with large appearance variations, \ie,~pose, expression, and occlusion from clutter background. On the other hand, by going deeper, the spatial information will lose through pooling or convolution operations. This spatial information is essential to recognize tiny objects.
This problem can be partially alleviated by using a dilation operation and reducing the number of pooling operations, which have been widely applied to other computer vision applications, \ie,~image segmentation~\cite{chen14semantic,long2015fully}. However, the computation will dramatically increase with high spatial resolution of feature maps in the network, making it difficult to meet practical runtime speed. 

Recent face detection methods typically follow the paradigm of Faster-RCNN~\cite{renNIPS15fasterrcnn} and Single Shot MultiBox Detector (SSD)~\cite{liu2016ssd}, the two popular frameworks for object detection. Faster-RCNN proposes anchor boxes of different sizes to deal with different scales.
The range of scale it can handle is limited by the granularity of the convolutional feature map output by the last convolutional layer shared among region proposal and detector networks.
SSD employs multi-scale deep features to jointly estimate for class probabilities and bounding box regression.
The multi-scale inference helps to detect objects of different scales but each of its stages is not specially trained to handle a specific scale range. In other words, no constraints are enforced during the training stage that a stage must do well on certain scales.

We believe that faces with different scales possess different inherent visual cues and thus lead to disparate detection difficulties, which can be more effectively modeled with different specialized network structures. Figure~\ref{fig:intro} shows the score distribution of faces in the scale of $10px-40px$ with background in the similar scale. 
This figure suggests that a specialized network, which needs not be deep, but with carefully designed depth and spatial pooling, can obtain very competitive results compared to state-of-the-art models when it focuses on a specific scale range.

To this end, we present a novel approach of designing a scale-friendly face detection deep network. It splits a large range of target scales into a set of sub-ranges. Each sub-range is modeled by a specialized network with carefully designed depth and spatial pooling to optimize the receptive field for the particular range. These networks can be seamlessly combined into a single network (\eg, resembling the structure of ResNet-50), and thus optimized in an end-to-end fashion. 
In the methodology section, we systematically explore and discuss different aspects of designing the network, including the way we find the best detection scale ranges for a network and the plausible configurations of combining multiple scale-specific detectors into a single unified network.

The proposed face detection network, dubbed ScaleFace, achieves promising performance on challenging benchmark datasets, namely, WIDER FACE~\cite{yang2016wider} and FDDB~\cite{fddb}, with practical runtime speed. WIDER FACE~\cite{yang2016wider} is the most challenging data given its wide spread of face scales. The best method to date of submission is~\cite{hu2016finding}, which achieves an average precision of $81\%$, and runs at 0.6 fps. Our method, \textit{ScaleFace},~achieves an average precision of \textbf{76.4} with just \textbf{7} fps.
The fundamental improvement in speed comes from two parts. Our method enjoys a more efficient way to integrating different scale-specific networks, which allows us to use a smaller backbone model while achieving good performance. Importantly, we use a single scale inference rather than image pyramid inference.

\section{Related Work}
\label{sec:related_work}

\noindent
\textbf{Deep learning based face detection}.
Following the remarkable performance of deep convolutional networks on image classification~\cite{russakovsky2015imagenet} and object detection~\cite{RCNN}, recent face detection studies~\cite{chen2016supervised,cascadecnn,li2016face3d,opitz16gridloss,yang2015faceness} also embrace deep learning for improved performance. These methods use deep convolutional networks as the backbone structure to learn highly discriminative representation from data and achieve impressive results on benchmark datasets such as FDDB and AFW. Among these methods, Faceness-Net~\cite{yang2015faceness}, STN~\cite{chen2016supervised}, and Grid-Loss~\cite{opitz16gridloss} are designed to detect faces under occlusions and large pose variations, Cascade-CNN~\cite{cascadecnn} and its variants~\cite{qin2016joint} achieve a good trade-off between speed and accuracy. Meanwhile, the unsatisfactory performance of existing methods on recent benchmark datasets in object detection~\cite{lin2014microsoft} and face detection~\cite{yang2016wider} reveals a new challenge on detecting tiny objects in uncontrolled environments.

There are unique and inherent challenges in multi-scale face detection that require special and systematic analysis. In this study, we are trying to detect faces in an extremely large range of scale, of which the variance is much larger than object detection, \ie, the target scale of object detection lies in [30-300]~\cite{liu2016ssd} while that of face detection is [10-1000]. In addition, tiny faces usually appear very close to each other in a crowded scene. The design of appropriate receptive fields for different face scales becomes essential.

\noindent
\textbf{Multi-scale face detection}.
A number of studies~\cite{zhang2010survey} have been proposed to address multi-scale face detection. Recent deep learning-based methods can be categorized into two classes: scale-invariant based methods~\cite{cascadecnn,jiang16fasterrcnn} and scale-variant based method~\cite{hu2016finding,liu2016ssd,qin2016joint}.

\noindent
(1) \textit{Scale-invariant methods}:
The vast majority of face detection pipelines focus on learning scale-invariant representation. The seminal work of Faster-RCNN~\cite{jiang16fasterrcnn} subscribes to this philosophy by extracting scale-invariant features through region of interest (ROI) pooling. The Cascade-CNN~\cite{cascadecnn} normalizes target object into a fixed scale and conducts multi-scale detection through an image pyramid.  However, Faster-RCNN and Cascade-CNN are not specifically designed to finding faces in a wide range of scales. 
Specifically, the foreground and background ROIs of Faster-RCNN map to the same location on deep features, causing ambiguity to the classifier. The Cascade-CNN is mainly formed by a set of three-layer CNNs thus its capacity confines it from handling large appearance and scale variances at the same time.

\begin{figure*}[t]
\begin{center}
{\includegraphics[width=0.95\linewidth]{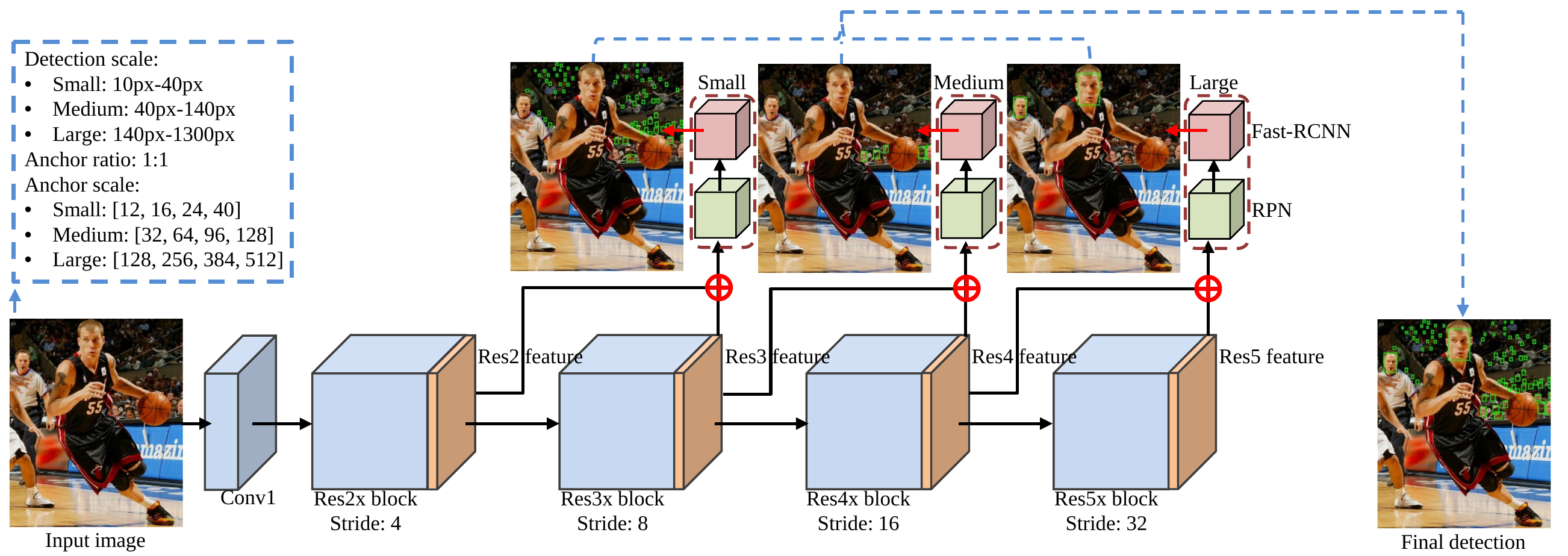}}
\vskip -0.1cm
{\caption{The pipeline of ScaleFace. The proposed network contains three scale-variant detectors which are incorporated into a single backbone network, which allows all detectors to be optimized end-to-end. Each scale-variant detector generates scale-specific detection predictions. The final detection results are obtained by aggregating all scale-variant detector results.}\label{fig:pipeline}}
\vspace{-0.7cm}
\end{center}
\end{figure*}

\noindent
(2) \textit{Scale-variant methods}: 
In contrast to learning scale-invariant representation, Qin~\etal~\cite{qin2016joint} propose a joint cascade network for learning scale-variant features. Samples from different scales are modeled separately by different networks and the detection results are generated by merging predictions across networks.  Similar to Cascade-CNN, the capacity of the individual network in the joint cascade network is insufficient to handle large scale and appearance variances. SSD~\cite{liu2016ssd} is proposed for object detection by making use of scale-variant templates based on the deep features. The SSD essentially tries to detect objects of various scales at different stage/layer of the network. 
Nevertheless, a direct application of SSD for small face detection still does not return satisfactory results (see Fig.~\ref{fig:intro}) since the scale-variant templates at early layers cannot cope well with large-scale variance. 
While in the later stages, SSD will suffer similar overlap mapping problem as in Faster-RCNN.

Recently, Hu~\etal~\cite{hu2016finding} propose a multi-task context-aware framework and achieve impressive results on detecting tiny faces. They use very deep neural networks and maintain spatial pooling stride by $8$ to alleviate the overlap mapping problem. In addition, features from different stages are aggregated to enhance the discriminative power for classification. A large set of scale-variant templates is learnt and multi-scale inference is required during testing. This method is computationally expensive. 

\noindent
\textbf{Differences with previous methods}.
Our method differs from previous methods in many aspects. We argue that given a deep neural network there exists a best range of scale for detection. Conceptually different from scale-invariant based methods, we address face detection through training specialized networks with the most suitable depth and spatial pooling stride, so that they are good in each specific sub-range of scales. 
Our method differs from SSD. The latter can be viewed as a cascade framework with each stage generating predictions to cover certain scales.
However, samples that cannot be detected in the early stage still need to be retrieved in the later stage. 
This effectively requires every stage to generalize well in dealing with large-scale variance.
In contrast, each stage in our framework is trained to detect faces fall within a certain range of scales. That allows each stage to specialize rather than generalize as in SSD.  
Different from Hu~\etal's method~\cite{hu2016finding}, which is formulated as a multi-task learning network, we incorporate a set of scale-variant networks into a single network for end-to-end optimization. 
This not only reduces computations through sharing parameters but also improves recognition performance by learning more discriminative representation jointly.

\section{Learning a Scale-Friendly Face Detector}
\label{sec:method}
In this section, we first provide an overview of our framework and then discuss the details. Figure~\ref{fig:pipeline} depicts the overall test pipeline of ScaleFace. 
Our framework contains three scale-variant detectors with different size of spatial pooling stride and depth. The scale-variant detectors are integrated into a single backbone network by sharing representation. 
The single backbone network turns out to have an identical structure as ResNet-50~\cite{he2016deep}. We extract features from the last layer of each res-block,~\ie, (res2cx, res3dx, res4fx, res5cx), from this backbone network. The features are used to train a region proposal network and a Fast-RCNN classifier. We henceforth refer to these features as (res2, res3, res4, res5) features. We use ROI pooling operation to extract features of different regions for detection.
With this unique structure that incorporates different scale-variant detectors, our algorithm can detect faces of different sizes efficiently by just using a \textit{single scale inference}, \ie, using a single input image without an image pyramid.

Given a test image, a forward pass is performed and each scale-variant face detector will generate detection windows independently. As shown in Figure~\ref{fig:pipeline}, each scale-variant face detector can detect faces fall within a certain range of scale. The detection windows generated from different scale-variant face detectors are merged and non-maximum suppression (NMS) is applied to eliminate highly overlapped windows.

In the following subsections, we present our considerations and insights in designing the network. We accompany our discussions with some experimental studies to gain a more detailed understanding. We follow the Scenario-Int protocol on WIDER FACE dataset and use $12,921$ images for training.
We report results on WIDER FACE validation set that contains $3,200$ images. We use the official evaluation toolbox~\cite{yang2016wider} to perform evaluation and use average precision (AP) to measure face detection performance.

To frame the discussion, we ask three questions. \textit{Given a network structure, what is the best range of scale for this network to perform detection?} We show that there exists an intrinsic correlation between network structure and detection scale. 
Since different scales can be modeled by different networks, the next question is, \textit{given a target range of scale, how many networks are sufficient to cover the whole target range?}
Suppose we have a set of face detectors with different network structures, we are interested about how to combine these convolutional networks into a single unified network to gain efficiency and performance. Namely, \textit{What is the best arrangement of the networks?}

\subsection{Finding a network for specific scale range}
\label{sec:scale-variant_classifier}

We hypothesize that faces with different scales can be better modeled by networks with different spatial pooling structure. Specifically, the projected face scale needs to match the ROI pooling size.   
To verify the effectiveness of this design, we first group faces into three classes according to the image height: small($10px-40px$), medium ($40px-140px$), and large($140px$ or more).
For each face group, we train four deep networks with different spatial pooling structure. Specifically, we train networks with a stride chosen from the set of $\left\{4,8,16,32\right\}$, which is based on the ResNet-50 structure shown in Fig.~\ref{fig:pipeline}. For example, the network with a stride of $4$ is constructed using layers from Conv1 to the Res2c. Multi-scale feature (``Hypercolumns''~\cite{hariharan2015hypercolumns}) from the current stage and previous stage (which has half number of stride of the current stage) are used and ROI pooling with a template size of $5\times5$ is applied for feature extraction as shown in the Fig.~\ref{fig:pipeline}.

\begin{table}[t]
\begin{center}
\caption{Average precision on WIDER FACE validation set. We evaluated different networks on faces of different scales.}
\vspace{0.1cm}
\label{tab:exp_best_structure}
\footnotesize{
\addtolength{\tabcolsep}{-1pt}
\begin{tabular}{c|c c c c}
\hline
\multirow{2}{*}{Face scale}& Conv1+Res2 & Res2+Res3 & Res3+Res4 & Res4+Res5\\
 & stride=$4$ & stride=$8$ & stride=$16$ & stride=$32$ \\
\hline\hline
Small& $56.21\%$& $\textbf{60.99\%}$& $54.51\%$ & $49.97\%$\\
\hline
Medium& $69.54\%$& $71.49\%$& $\textbf{74.54\%}$ & $69.33\%$\\
\hline
Large& $58.43\%$& $72.19\%$& $81.89\%$ & $\textbf{84.68\%}$\\
\hline
\end{tabular}
}
\end{center}
\end{table}

\begin{figure}[t]
\begin{center}
\includegraphics[width=0.95\linewidth]{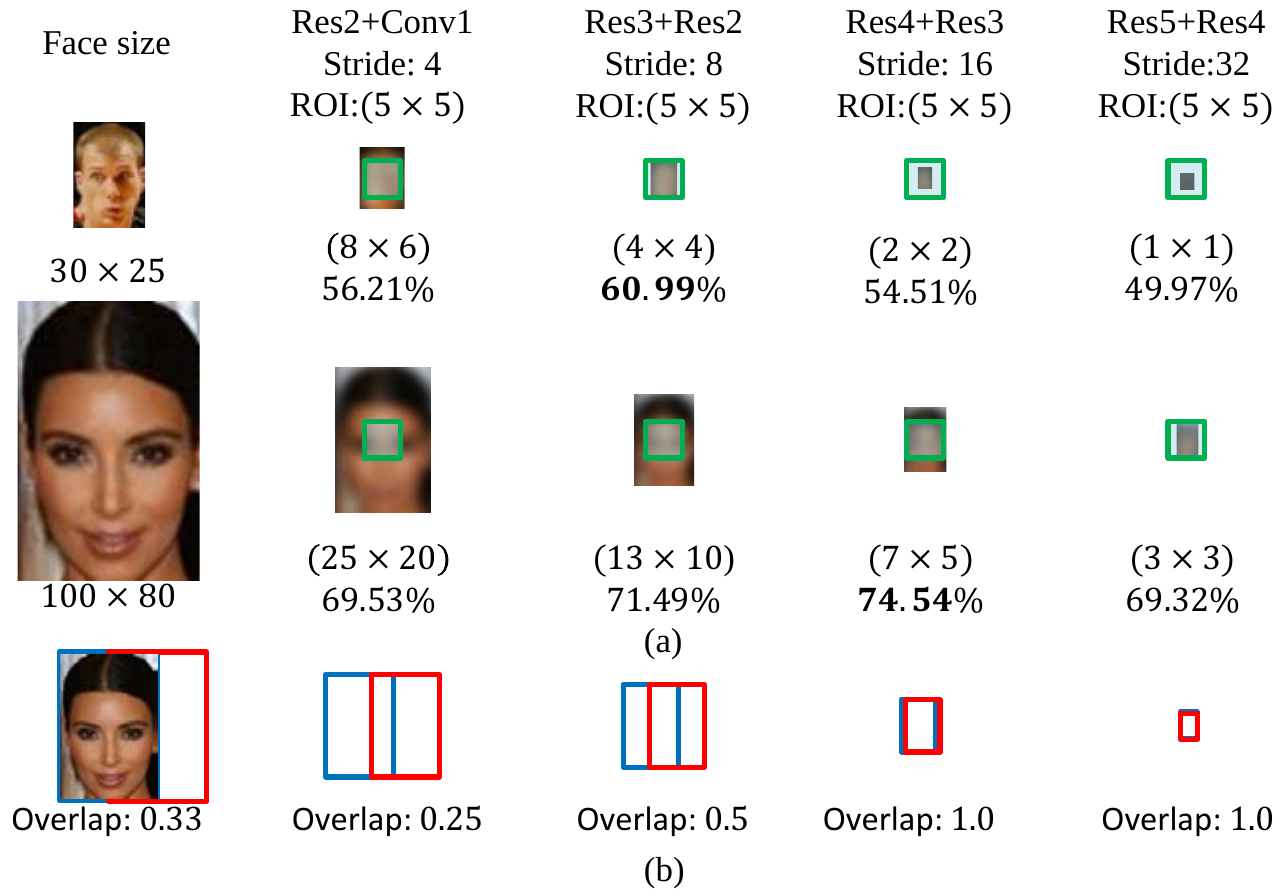}
\vskip -0.1cm
\caption{(a) The average precision of different network structure for different scale of faces. The green box represents the ROI template. (b) The degree of overlap between foreground and background receptive fields at different stages of a deep network. The blue box represents foreground while the red box represents background.}
\vspace{-0.7cm}
\label{fig:right_scale}
\end{center}
\end{figure}

Table~\ref{tab:exp_best_structure} summarizes the average precision of various possibilities on WIDER FACE validation set. The small faces ($10px-40px$) achieve the best performance using a network with a stride of $8$. The projected face scale on the feature map is $2px-5px$, which is in the similar scale of ROI template. Similarly, we observe the best performance of certain scales when the projected face scale on the feature map is close to the ROI template.
This observation is consistent across three face groups. 
We investigate further by studying Fig.~\ref{fig:right_scale}(a), which shows the influence of different network structures towards the detection performance of faces with different sizes. 
Recall that, for a proposed ROI, the size of ROI on the target layer is determined by the spatial pooling parameters. Typically convolutional features at higher layers tend to have smaller projected ROI size.  As we can observe from Figure~\ref{fig:right_scale}(a), the detection performance of a target scale consistently decreases when the ROI on the target layer is smaller than ROI pooling size. Even if we increase the depth of the network which will generally improve the discriminative power of the feature representation, the detection performance still drops. We verify that this is due to overlapping between feature maps after spatial transformation. Specifically, as shown in Figure~\ref{fig:right_scale}(b), the background region will map to the same location of the foreground region which will introduce ambiguity for the classifier.

From Figure~\ref{fig:right_scale}(a), we observe that using multi-scale feature (known as ``Hypercolumns''~\cite{hariharan2015hypercolumns}) alone is not sufficient to prevent performance drop. Remapped features with a similar size of ROI template will yield the best detection performance. If the projected region is much larger than the ROI template discriminative information will loss during pooling procedure. On the other hand, if the projected region is much smaller than the ROI template, the insufficient information and overlapping between features will cause a performance drop. 
The results in Table~\ref{tab:exp_best_structure} confirm our observation that samples with different scale need to be modelled with different network structures, \ie,~small face ($10px$-$40px$) should be modeled using network with spatial a stride of $8$ while larger face ($>200px$) can use a network with a stride of $32$. 

This experiment shows that the spatial resolution of a network is the key factor to achieve good detection performance for samples falls within a certain range of scales.

\subsection{How many scale-variant detectors}
\label{sec:split_range}
Given the observations in Sec.~\ref{sec:scale-variant_classifier}, we now discuss another question. If our target object lies on a very wide range of scales, how can we split the target range into a set of sub-ranges to achieve good detection performance?  Since different network structures would have different face detection performances, the trivial solution will be training a scale-variant detector for every single scale. If the target range is $[n,m]$, where $n<<m$, there will be $m-n+1$ number of classifiers. 
This approach would result in a number of highly redundant networks.
To another extreme, we can train a single network to cover the whole target scale, which is exactly the scale-invariant based method. This type of network structure is not optimal to cover the full target scale, as discussed in Sec.~\ref{sec:related_work}.

We hope to find a good trade-off between accuracy and runtime speed. A straightforward way is to split the target range uniformly into $k$ portions and adopt the strategy discussed in Sec.~\ref{sec:scale-variant_classifier} to obtain the best network for each sub-range. However, it may not be a viable solution since appearance variation is not uniformly distributed along the scale. 
Specifically, from our observation, small faces (less than $40$ pixel height) lose most appearance information and can be characterized by rigid structures and context. Medium faces ($40px-140px$) have high variance since persons in these images are usually not the main subjects of the photographer, and therefore they can be of various poses looking at different directions. Large faces ($140px$ or more) usually have low variance as they are the main subjects when a photo is captured. These large faces are usually in a frontal or profile pose.
This simple observation provides us with a clue on partitioning the scale range. When this observation is considered collectively with network selection of suitable spatial pooling stride, a huge gain can be obtained in comparison to a uniform partition of range.    

To validate our hypothesis, we perform an experiment to compare the uniform partition of range (`even splits') and the partitioning scheme inspired by our observation, which considers both appearance variation and suitable spatial pooling stride.
Table.~\ref{tab:exp_split} shows the range of scales using different split criteria.  
Again, we use ResNet-50 as the backbone network to conduct this experiment. Different target scales are assigned to different stages of the backbone network with suitable spatial pooling stride as shown in Table.~\ref{tab:exp_split_fea}. 
The average precisions of face detection on WIDER FACE validation set are shown in Table.~\ref{tab:exp_num_split}. The `three splits' network outperforms other methods consistently under three evaluation settings. `One split' network and `two splits' network report inferior performance on detecting small faces, since small faces $[10px-40px]$ are modeled using Res4 and Res5 features with pooling stride $16$ and $32$, respectively. The four splits network fails to detect small faces because of insufficient discriminative power of Res2 features to distinguish small faces from clutter background. 
The partitioning scheme based on appearance variation outperforms the evenly split baselines across all different number of splits. The main reason is that evenly split scheme does not take network structure into consideration. A miss correspondence between detection scale and network structure would jeopardise the detection performance. 

This experiment shows that partitioning the scale range based on face variation, and at the same time considering network selection of suitable spatial pooling stride, would give us useful clues on how many scale-variant detectors we should build.

\begin{table}[t]
\begin{center}
\caption{The range of scales under different range partitioning schemes.} \vspace{0.1cm}
\label{tab:exp_split}
\footnotesize{
\addtolength{\tabcolsep}{-1pt}
\begin{tabular}{c|c}
\hline
Method & Split ranges
\\
\hline\hline
One split & $[10,1300]$\\
\hline
Two splits& $[10,140]$, $[140,1300]$\\
Two evenly splits& $[10,650]$, $[650,1300]$\\
\hline
Three splits& $[10,40]$, $[40,140]$, $[140,1300]$,\\
Three evenly splits& $[10,450]$, $[450,900]$, $[900,1300]$,\\
\hline
Four splits& $[10,25]$, $[25,60]$, $[60,140]$, $[140,1300]$\\
Four evenly splits& $[10,300]$, $[300,600]$, $[600,900]$, $[900,1300]$\\
\hline
\end{tabular}
}
\end{center}
\end{table}

\begin{table}[t]
\begin{center}
\caption{The assignment of network stages to different splits.}
\label{tab:exp_split_fea}
\footnotesize{
\addtolength{\tabcolsep}{-1pt}
\begin{tabular}{c|c c c c}
\hline
\multirow{2}{*}{} & Conv1+Res2 & Res2+Res3 & Res3+Res4 & Res4+Res5 \\
 & stride=$4$ & stride=$8$ & stride=$16$ & stride=$32$ \\
\hline\hline
One split & & & & \Checkmark \\
Two splits & & & \Checkmark & \Checkmark \\
Three splits & & \Checkmark & \Checkmark & \Checkmark \\
Four splits & \Checkmark & \Checkmark & \Checkmark & \Checkmark \\
\hline
\end{tabular}
}
\end{center}
\end{table}

\begin{table}[t]
\begin{center}
\caption{Evaluation of different range partitioning schemes across three difficulty settings of WIDER FACE (Easy, Medium, Hard).} \vspace{0.1cm}
\label{tab:exp_num_split}
\footnotesize{
\addtolength{\tabcolsep}{-1pt}
\begin{tabular}{c|c c c}
\hline
Method & Easy & Medium & Hard
\\
\hline\hline
One split& $82.4\%$& $79.3\%$& $62.4\%$\\
Two splits& $83.0\%$& $83.5\%$& $74.7\%$\\
Three splits& $\textbf{86.8\%}$& $\textbf{86.7\%}$& $\textbf{77.2\%}$\\
Four splits& $84.2\%$& $85.1\%$& $72.1\%$\\
\hline
Two evenly splits& $78.4\%$& $79.5\%$& $62.8\%$\\
Three evenly splits& $72.2\%$& $73.6\%$& $57.1\%$\\
Four evenly splits& $68.4\%$& $69.0\%$& $53.1\%$\\
\hline
\end{tabular}
}
\end{center}
\end{table}

\begin{figure*}[t]
\begin{center}
\includegraphics[width=0.9\linewidth]{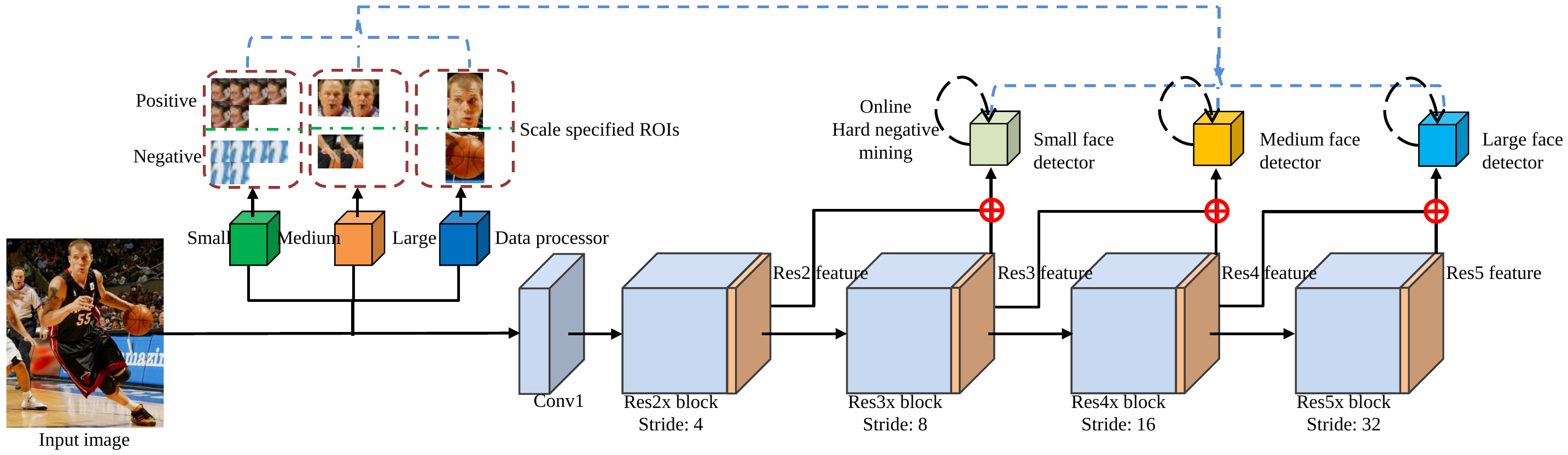}
\vskip -0.2cm
\caption{\small{The data processing procedure during the training stage of ScaleFace.}}
\vskip -0.3cm
\label{fig:data_processor}
\end{center}
\vskip -0.3cm
\end{figure*}

\subsection{How to combine the scale-variant detectors}
\label{sec:ensemble_classifier}
It is natural to ask a follow-up question: how to combine a set of scale-variant detectors. Previous face detection studies~\cite{qin2016joint,yang2016wider} train independent detectors and aggregate predictions generated from these detectors. There exists high redundancy between the detectors.
As we have shown in Sec.~\ref{sec:scale-variant_classifier}, different face scales demand networks of different structures. We can actually share representation among these networks to reduce redundancy.

The proposed network shown in Fig.~\ref{fig:pipeline} shares representations between the three scale-variant detectors. We compare it with a straightforward method that trains scale-variant detectors independently, using the similar structures as our detectors but without representation sharing. The predictions generated from each scale-variant detectors are aggregated to generate the final predictions. We call this method `na\"{i}ve ensemble' and our method as `joint optimized'. We select top performed scale-variant detectors based on the results shown in the Sec.~\ref{sec:split_range} from each scale ranges to conduct this experiment. Both `na\"{i}ve ensemble' and `joint optimized' use the same set of detectors.  The evaluation results on the WIDER FACE validation set are shown in Table.~\ref{tab:exp_ensemble}. The `joint optimized' method outperforms the `na\"{i}ve ensemble' method consistently across the three WIDER FACE evaluation settings.

This experiment suggests that we should try best to share representation between scale-variant detectors for joint optimization.

\begin{table}[t]
\begin{center}
\caption{Evaluation of ensembles. We report average precision of `na\"{i}ve ensemble' and `joint optimized' on the WIDER FACE validation set.}
\label{tab:exp_ensemble} \vspace{0.1cm}
\footnotesize{
\addtolength{\tabcolsep}{-1pt}
\begin{tabular}{c|c c c}
\hline
Method & Easy & Medium & Hard
\\
\hline\hline
Navies ensemble & $73.2\%$& $74.1\%$& $60.8\%$\\
Joint optimize& $\textbf{86.8\%}$& $\textbf{86.7\%}$& $\textbf{77.2\%}$\\
\hline
\end{tabular}
}
\end{center}
\vskip -0.4cm
\end{table}

\subsection{Training and implementation details}
\label{sec:implementations}

During the training of ScaleFace, each detector has its own disjoint sets of training samples as shown in the Figure~\ref{fig:data_processor}, which samples scale-variant regions from the input image. Each detector should only sample scales that belong to the range it should handle. For example, the detector targeting at $10px$-$40px$ faces, will be trained with positive and negative regions in the range of $10px$-$40px$, regions excluded from this range will not be sampled. 
Consequently, different scale-variant detectors are optimized using different ROIs and scale-variant ground-truth. 
The ROIs are generated by comparing intersection overlap union (IOU) between ground-truth across scales. The ground-truth regions out of the scale range a detector should handle are assigned as an `ignored' class. Those samples that with high IOU but belong to the ignored class will still be discarded. Faces of particular scales may be under-represented in the data. In such cases, we will balance the data during the training process.

We train and test our framework on images of a single scale. The images are rescaled to keep the longer side no more than $1,300$. As shown in Figure~\ref{fig:pipeline}, each scale-variant detector has no more than $4$ anchors with an aspect ratio of $1:1$. We do not tune these hyper-parameters for a particular dataset. Samples with IOU over $0.5$ are assigned as positive samples. The negative samples originate from two sources -- we sample regions with IOU in the range of $0.0-0.1$ and $0.1-0.3$ in a balanced manner since both pure background regions and ill-aligned regions are important. Specifically, we construct background pool by randomly sample the same number of pure background regions with the number of ill-aligned regions, therefore, the number of pure background regions and ill-aligned regions is $1:1$ in the background pool. Then we uniformly sample background regions from this pool. On-line hard negative mining for training Fast-RCNN branches are used. We adopt non-maximum suppression (NMS) on the proposal regions generated by RPN and keep top $2,000$ proposals during training and $500$ proposal regions during testing. Predictions generated from scale-variant classifiers are aggregated and NMS with a threshold of $0.3$ is adopted to generate the final results. 

\section{Model compression}
\label{sec:model_compression}

To improve the runtime efficiency of our face detector, we conduct model compression to reduce parameters. Specifically, we reduce the number of filters in every layer of the backbone network and pretrain this new network on ImageNet dataset on the $1,000$-objects recognition task. Next, we finetune the compressed model on the face detection task using the same procedure introduced in Sec.~\ref{sec:ensemble_classifier}.      

\section{Results on benchmark datasets}
\label{sec:experiments}

\noindent \textbf{Training datasets}.
We use WIDER FACE dataset~\cite{yang2016wider}, which is proposed based on WIDER dataset~\cite{xiongevent15}, to train our model. The WIDER FACE dataset contains $32,203$ images and $393,703$ annotated faces. We follow the Scenario-Int protocol on WIDER FACE dataset and use $12,921$ images for training. As mentioned in the Sec.~\ref{sec:implementations}, data augmentation is necessary to train our face detector. Images that contain more than $25$ valid faces are duplicated $5$ times as additional data, since most faces in the WIDER FACE dataset appear in these $20\%$ of images. The data augmentation can ensure the network has enough images used for learning small faces. Horizontal flipping is applied on all training images. The augmented training set contains $44,000$ images.

\noindent \textbf{Test datasets}.
We verify the effectiveness of ScaleFace on two benchmark datasets: WIDER FACE~\cite{yang2016wider} and FDDB~\cite{fddb}. Since we use WIDER FACE training split to train our face detector, we follow Scenario-Int evaluation protocol and test on the WIDER FACE test split. The WIDER FACE test set contains $16,000$ images with $196,850$ annotated faces. The FDDB dataset contains annotations for $5,171$ faces in a set of $2,845$ images. We follow the standard evaluation protocol on these datasets -- average precision (AP) is used as evaluation metric on the WIDER FACE dataset and ROC evaluation is used on the FDDB dataset.

\subsection{Evaluate on benchmark datasets}
\label{sec:exp_benchmark}

We evaluate our face detector against state-of-the-art face detection methods on FDDB~\cite{fddb} and WIDER FACE~\cite{yang2016wider} datasets.  

\begin{figure}[t]
\begin{center}
\includegraphics[width=1\linewidth]{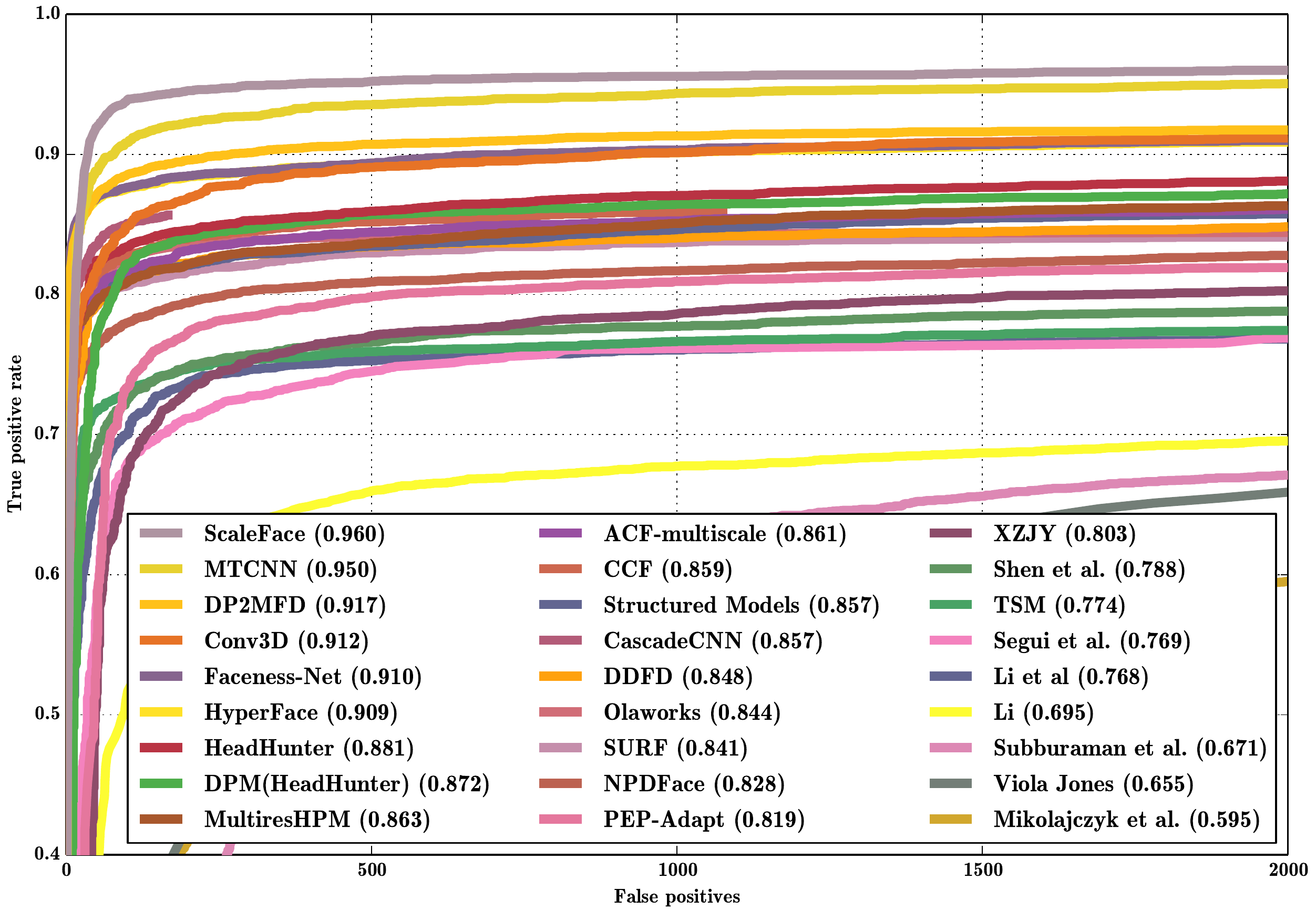}
\caption{\small{FDDB results. Recall rate is shown in the parenthesis.}}
\label{fig:fddb_dis}
\end{center}
\vspace{-0.5cm}
\end{figure}

\noindent
\textbf{FDDB}: 
We test our model trained on WIDER FACE directly on FDDB without fine-tuning. There exists a mismatch between annotations in WIDER FACE and FDDB. In particular,  the former uses bounding box annotation and the latter employs bounding ellipse annotation. Consequently, we follow the standard practice~\cite{HeadHunter} and train a linear regressor to transform our predicted bounding boxes to meet FDDB annotation style. We compare our method with state-of-the-art face detection methods~\cite{JointCascade, HeadHunter, yang2015faceness, ranjan16hyperface, zhang16jointaligment, li2016face3d}. The results on discrete score are shown in Fig.~\ref{fig:fddb_dis}. Our method achieves $94.55\%$ recall rate with $200$ false positives and $96\%$ recall rate with $2,000$ false positives, outperforming the best baseline method by $2.3\%$ and $1.0\%$ recall rate, respectively. The result demonstrates the effectiveness and good generalization capability of ScaleFace.  

\begin{figure}[t]
\begin{center}
{\includegraphics[width=0.9\linewidth]{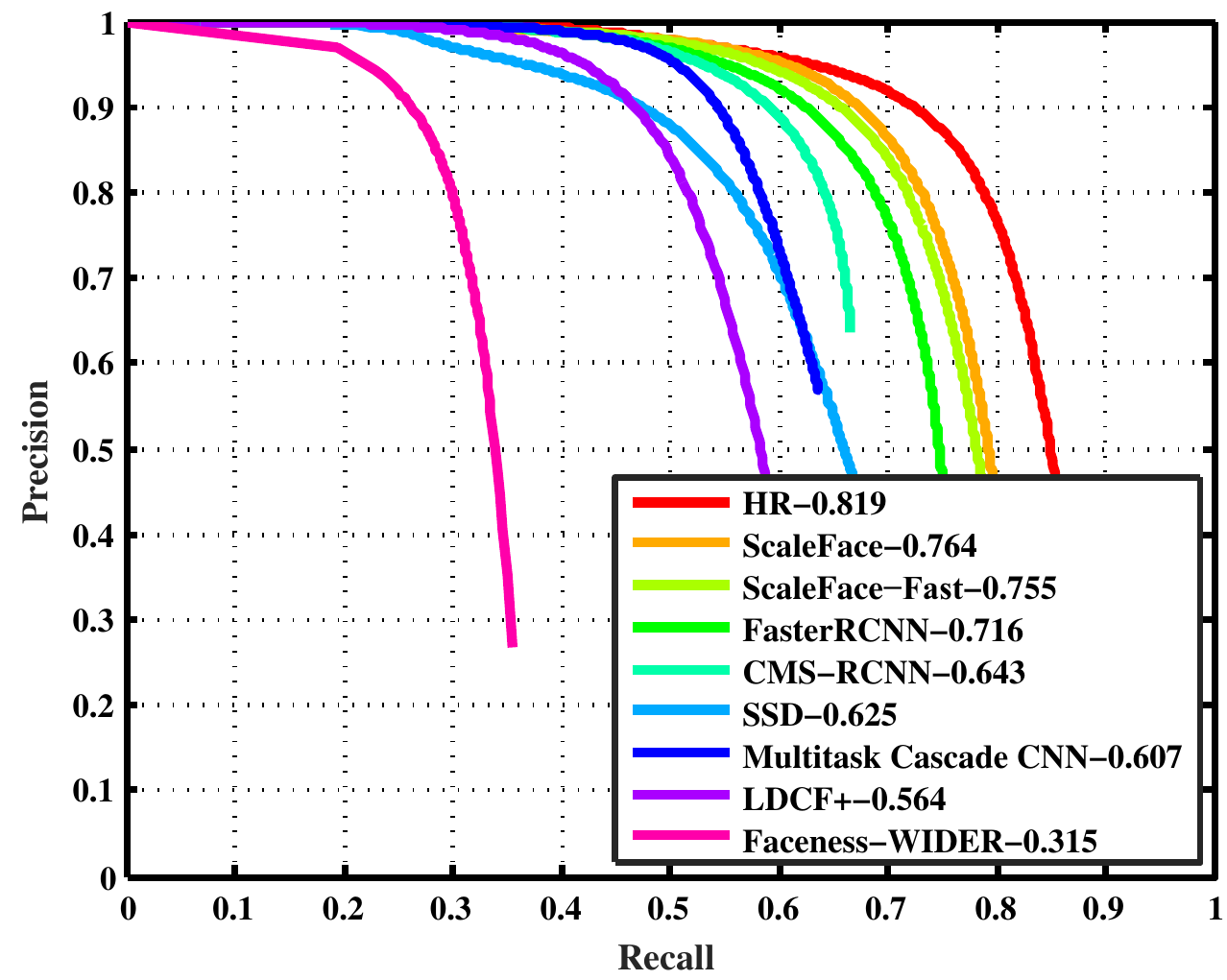}}
{\caption{Precision and recall curves on WIDER FACE hard set.}\label{fig:exp_wider_face}}
\vspace{-0.5cm}
\end{center}
\end{figure}

\begin{figure*}[t]
\begin{center}
\includegraphics[width=1\linewidth]{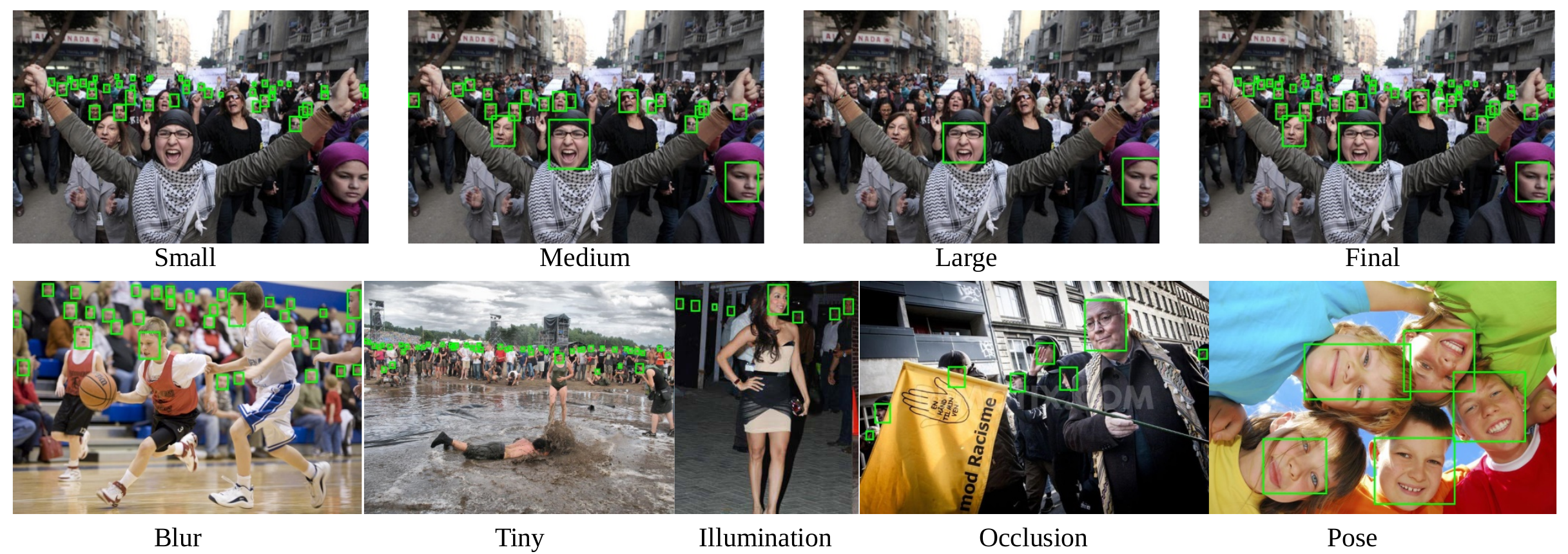}
\caption{\small{The first row shows the output of each stage in our network. The second row shows qualitative results with different attributes. More examples can be found in the appendix.}}
\label{fig:visual_sample}
\end{center}
\vspace{-0.2cm}
\end{figure*}

\vspace{0.1cm}
\noindent
\textbf{WIDER FACE}:
We train our face detector on WIDER FACE training set and report face detection performance on the held-out test set. The WIDER FACE dataset has three evaluation settings: easy, medium, and hard. It is observed that the difficulty settings correlate with the face scales. 
In the hard setting, all valid face annotations (faces with a height larger than $10px$) are used for evaluation. 
We compare our method with top performers~\cite{yang2017facenesspami,zhang16jointaligment,zhu16cms,hu2016finding,ohn2017boost} on WIDER FACE dataset. 
In order to verify the effectiveness of our method, we prepare two strong baselines, Faster-RCNN and SSD, which are built using ResNet-50. The baselines thus have the same model capacity as ScaleFace.  
We change the convolution stride in Res4f to $1$, which makes the maximum spatial stride of the network equals to $16$. The Faster-RCNN uses Res5 and Res3 features for classification and it is trained on the WIDER FACE training set with multi-scale data augmentation following the protocol presented in~\cite{hu2016finding}. For each stage of SSD, we apply the same strategy employed by our method, \ie, it also uses multi-scale features from the neighboring stage for the recognition.

Figure~\ref{fig:exp_wider_face} shows the precision and recall (PR) curves of different methods on WIDER FACE hard set. Our method, in general, shows a compelling result in the hard setting and outperforms FasterRCNN and SSD by a considerable margin ($4.8\%$ and $13.9\%$ respectively). Table~\ref{tab:exp_wider} summarizes the average precision on three evaluation settings.
The HR~\cite{hu2016finding} outperforms our method by using ResNet-101 with a stride pooling equals to $8$. A total of 25 scale-variant templates is used to generate detection results while multi-scale inference is required. 
It addition, HR learns additional post-regressors to improve face localization. Without this post-processing, the average precision on the hard set drops to $79.8\%$. If we switch the ResNet-101 with ResNet-50 as the backbone network of HR, its performance will drop by $2\%$ across three evaluation settings. Our method does not conduct post regression and achieves $76.4\%$ AP on WIDER FACE hard set, which is on par with HR using the same backbone network (ResNet-50) and without post-processing. 
It is worth pointing out that our method achieves state-of-the-art performance when the precision is over $95\%$ on the WIDER FACE hard set.

\begin{table}[t]
\begin{center}
\caption{Evaluation of different range partitioning schemes across three difficulty settings of WIDER FACE (Easy, Medium, Hard).} \vspace{0.1cm}
\label{tab:exp_wider}
\footnotesize{
\addtolength{\tabcolsep}{-1pt}
\begin{tabular}{c|c c c}
\hline
Method & Easy & Medium & Hard
\\
\hline\hline
Faceness-Net~\cite{yang2015faceness}& $71.6\%$& $60.4\%$& $31.5\%$\\
LDCF+~\cite{ohn2017boost}& $79.7\%$& $77.2\%$& $56.4\%$\\
MTCNN~\cite{zhang16jointaligment}& $85.1\%$& $82.0\%$& $60.7\%$\\
CMS-RCNN~\cite{zhu16cms}& $90.2\%$& $87.4\%$& $64.3\%$\\
HR~\cite{hu2016finding}& $92.3\%$& $91.0\%$& $81.9\%$\\
\hline
SSD~\cite{liu2016ssd}& $89.9\%$& $85.4\%$& $62.5\%$\\
FasterRCNN~\cite{renNIPS15fasterrcnn}& $89.5\%$& $87.1\%$& $71.6\%$\\
ScaleFace& $86.7\%$& $86.6\%$& $76.4\%$\\
\hline
\end{tabular}
}
\end{center}
\vspace{-0.6cm}
\end{table}

\subsection{Runtime analysis}
\label{sec:runtime_analysis}
As discussed in Sec.~\ref{sec:model_compression}, ScaleFace can be compressed by reducing the number of filters in every layer of the backbone network. In this experiment, we build a compressed model based on this method. First, we reduce the number of filters in every layer of ResNet-50 to half the original. We call this model as ScaleFace-Fast. The model is first pretrained on the ImageNet dataset and fine-tuned for face detection using the WIDER FACE dataset.

We compare ScaleFace and its compressed version with Faster-RCNN, SSD, and HR. All methods are tested using NVIDIA Titan X GPU by averaging the runtime of $1,000$ images randomly sampled from the WIDER FACE dataset. 
ScaleFace achieves $76.6\%$ average precision with $4 fps$ for $900\times1300$ image. Faster-RCNN and SSD record $7.1$ fps and $9.1$ fps for $900\times1300$ input but the average precisions are $4.8\%$ and $13.9\%$ lower than ScaleFace. HR achieves the best face detection performance, but it uses a very deep neural network with high-resolution feature maps. Dense scale templates are applied and multi-scale inference is required, which dramatically increase its computation. The runtime speed of HR is $0.6$ fps. ScaleFace runs $6$ times faster than HR.

Our compressed model ScaleFace-Fast achieves $75.5$ average precision on WIDER FACE test set. The ScaleFace-Fast needs only $160$ms for detecting faces in a $900\times1300$ image. The fast version of our method only experiences a small performance drop $0.9\%$, compared with the original ScaleFace. The compressed model runs $\textbf{10}$ times faster than HR, while achieves reasonably good detection performance with practical runtime speed.  


\begin{table}[t]
\begin{center}
\caption{Runtime analysis. The time is averaged across $1,000$ images with a resolution of $900\times1300$.}
\label{tab:exp_runtime}
\footnotesize{
\addtolength{\tabcolsep}{-1pt}
\begin{tabular}{c|c c c}
\hline
Method & AP & Runtime (ms)
\\
\hline\hline
FastRCNN & $71.2\%$& $140$\\
SSD & $62.4\%$& $110$ \\
HR & $81.9\%$& $1,600$\\
ScaleFace & $76.4\% $& $270$\\
ScaleFace-Fast& $75.5\%$ & $160$\\
\hline
\end{tabular}
}
\end{center}
\vspace{-0.6cm}
\end{table}

\section{Conclusion}
\label{sec:discussion}
In this work, we have proposed a simple yet effective framework to detect faces with very large-scale variance. Our method exhibits a good trade-off between performance and speed for face detection. We have demonstrated that both network structure and target scale are correlated. Understanding such correlation is a key to achieving good detection performance. Specifically, we show that different scale range needs to be modeled by different network structure. The size of target object after remapping to the feature maps used for recognition should be in the similar scale with ROI pooling feature. Given a very large range of scale, we can split the original scale into several sub-scales by considering appearance variations of faces. Finally, different scale-variant networks can be incorporated into a single network for end-to-end optimization. The ensemble can be further compressed without losing much performance while greatly improving its efficiency.

\appendix
\section{Appendix}
\subsection{Network structure}
\label{sec:network_structure}
\begin{figure*}[t]
\begin{center}
\includegraphics[width=\linewidth]{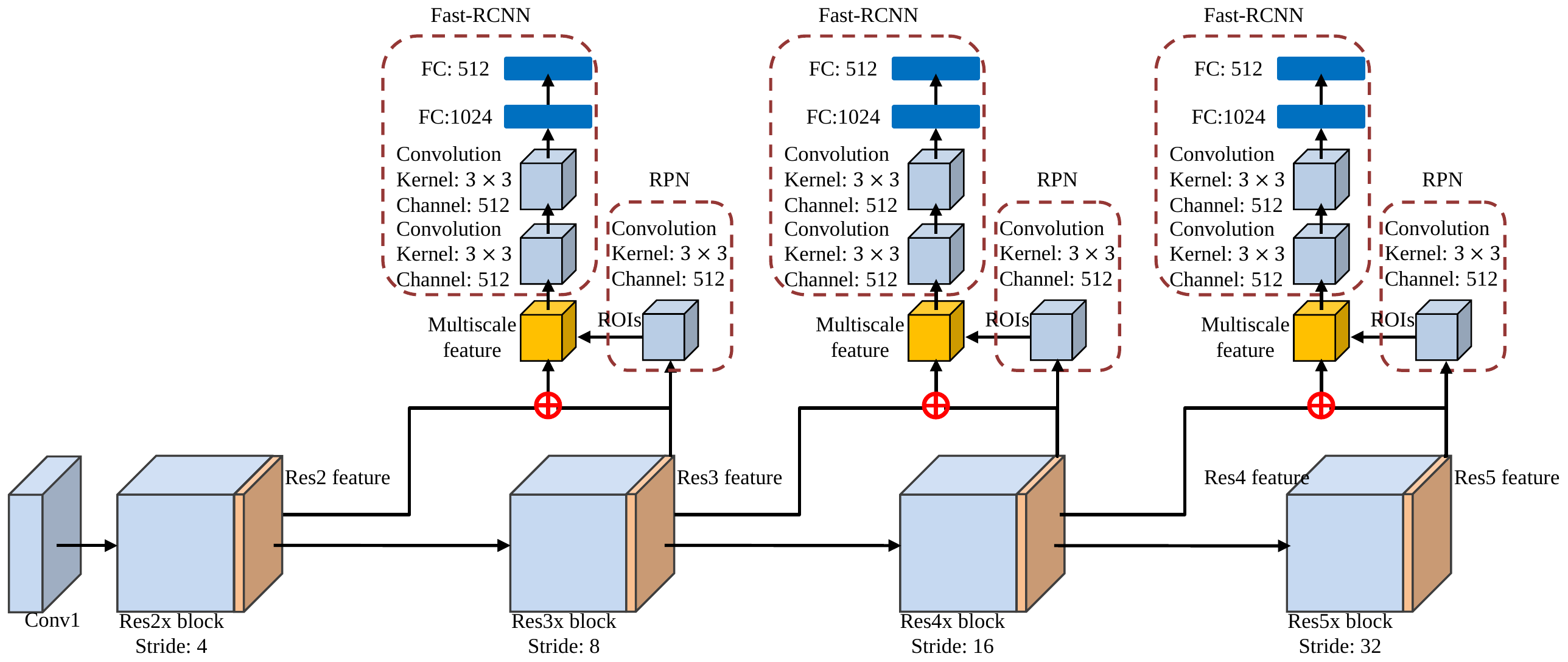}
\caption{\small{The detailed network structure of ScaleFace.}}
\label{fig:network}
\end{center}
\vspace{-0.2cm}
\end{figure*}

Figure.~\ref{fig:network} shows detailed network structure of our model. Each scale-variant detector consists of a region proposal network and a Fast RCNN classifier. The region proposal network is implemented by using a convolutional layer with $3\times3$ kernel and $512$ channels. The Fast RCNN contains two convolutional layers, each of which uses $3\times3$ kernel and $512$ channels and two fully connected layers with $1024$ and $512$ neurons as shown in Figure.~\ref{fig:network}.

\subsection{Visual samples of ScaleFace}
\label{sec:scale_face}

\begin{figure*}[t]
\begin{center}
\includegraphics[width=\linewidth]{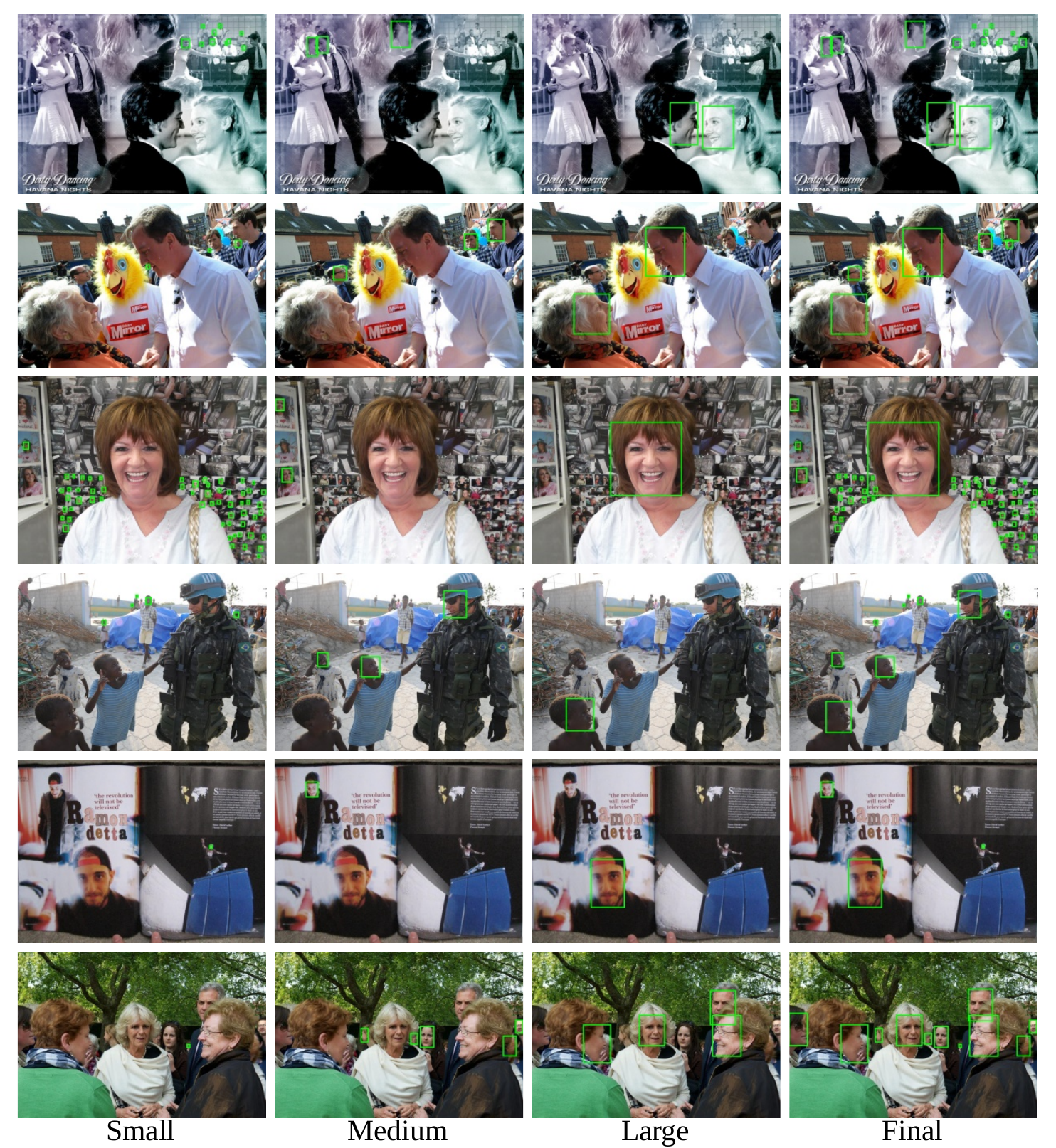}
\caption{\small{The output of each stage in our network.}}
\label{fig:visual_sample_scaleface}
\end{center}
\end{figure*}

Figure.~\ref{fig:visual_sample_scaleface} shows the predictions generated from different scale-variant detectors and final detection results. The scale-variant detectors only need to generate detection results in a certain range of scale. As we can observe from Figure.~\ref{fig:visual_sample_scaleface} the predictions generated by small face detector has no overlap with predictions generated by the large face detector. The medium face detector serves as a bridge to generate intermediate scale predictions with some overlaps with the small face detector and the large face detector to ensure high recall rate.  

\subsection{Compare with baseline methods}
\label{sec:compare}

\begin{figure*}[t]
\begin{center}
\includegraphics[width=0.8\linewidth]{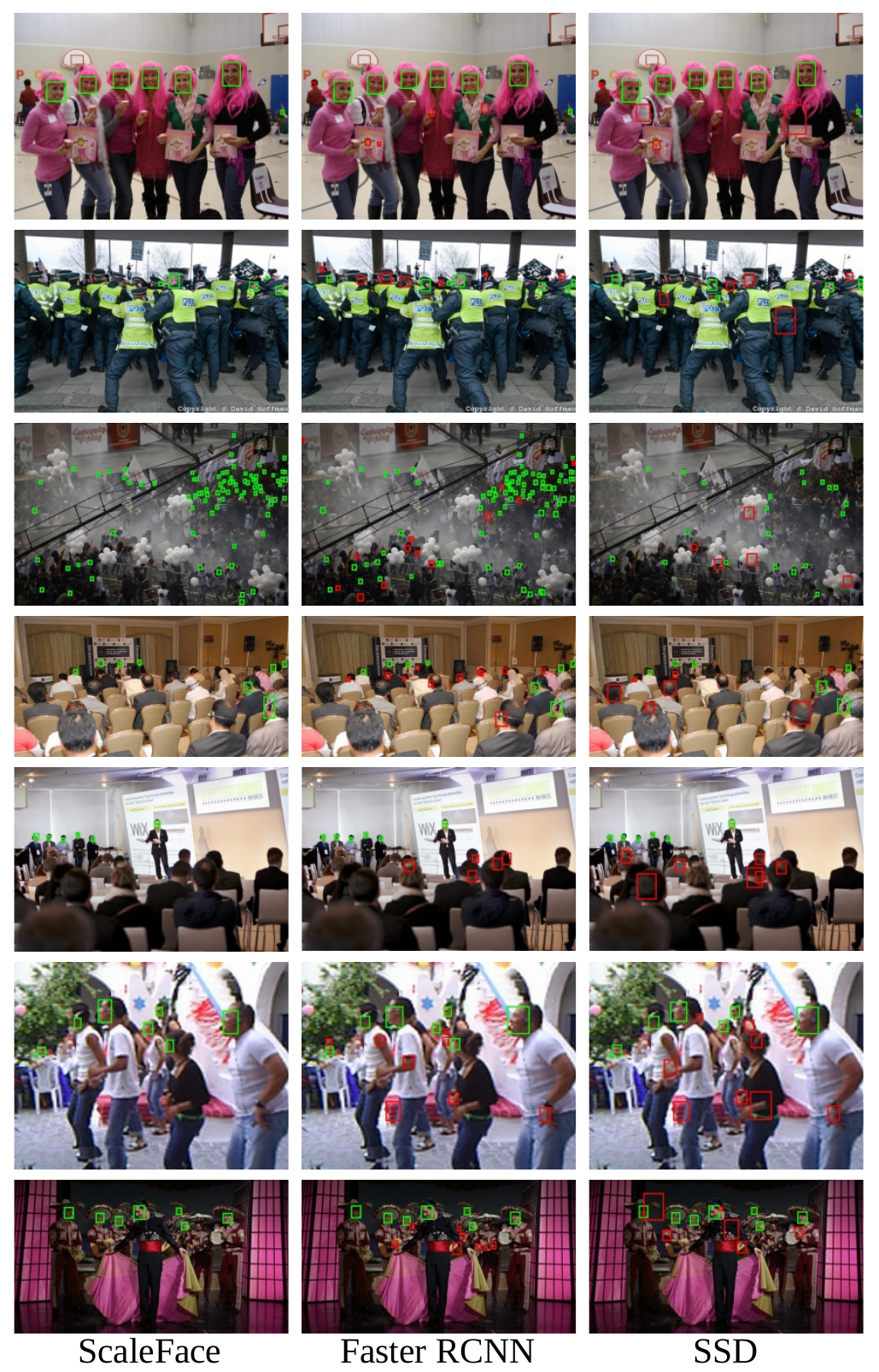}
\caption{\small{The results generated by ScaleFace, Faster RCNN~\cite{renNIPS15fasterrcnn}, and SSD~\cite{liu2016ssd}.}}
\label{fig:visual_sample_compare}
\end{center}
\vspace{-0.2cm}
\end{figure*}

We compare Faster-RCNN~\cite{renNIPS15fasterrcnn} and SSD~\cite{liu2016ssd} with the proposed ScaleFace. Figure~\ref{fig:visual_sample_compare} shows the results generated by different algorithms. Faster-RCNN generates a number of false alarms (indicated by red bounding boxes) on small-scale predictions. This is because projected ROIs for small faces at the deep layer quickly shrink to $1\times1$, which is much smaller than the ROI pooling size. This projection maps backgrounds and objects to a small region on the feature map and therefore introduces ambiguity to the classifier. SSD tries to make predictions on full target detection scale in every stage, this strategy works fine when objects do not have very large variance in scale. In the early stage of SSD, the network tends to generate false alarms because of limited representative power. In the later stage, it seems that the network overfits to medium and large faces. In contrast, our method consistently performs well across all scales.  

\newpage

{\small
\bibliographystyle{ieee}
\bibliography{syang_bib}
}

\end{document}